\documentclass[letterpaper, 10 pt, conference]{ieeeconf}

\IEEEoverridecommandlockouts                              
\usepackage{times}
\usepackage{multicol}
\usepackage{url}
\usepackage{color,soul}

\usepackage[dvipsnames]{xcolor}
\usepackage{subcaption}
\usepackage{amsmath,amssymb}

\usepackage{amsthm}
\usepackage{graphicx}  
\usepackage{tabularx, booktabs}

\usepackage[colorinlistoftodos]{todonotes}
\newcolumntype{C}{>{\centering\arraybackslash}X} 

\newtheoremstyle{myplain}
    {}
	{}
	{\itshape}
	{}
	{\bfseries}
	{}
	{5pt plus 1pt minus 1pt}
	{}

\newtheoremstyle{mydefinition}
	{}
	{}
	{\normalfont}
	{}
	{\bfseries}
	{}
	{5pt plus 1pt minus 1pt}
	{}
	
\theoremstyle{myplain}

\let\labelindent\relax
\usepackage{enumitem}

\usepackage[bookmarks=true]{hyperref}
\hypersetup{
    colorlinks=true,
    linkcolor=black,
    filecolor=black,
    urlcolor=magenta,
    pdftitle={HIO-SDF: Hierarchical Incremental Online Signed Distance Fields},
}

\newcommand{\ourmethod}{HIO-SDF}

\title{\LARGE \bf
\ourmethod{}: Hierarchical Incremental Online Signed Distance Fields
}

\author{Vasileios Vasilopoulos, Suveer Garg, Jinwook Huh, Bhoram Lee and Volkan Isler
\thanks{Work performed while at Samsung AI Center NY.}%
}

\renewcommand{\baselinestretch}{.96}

\begin{document}

\maketitle
\thispagestyle{empty}
\pagestyle{empty}

\begin{abstract}
A good representation of a large, complex mobile robot workspace must be space-efficient yet capable of encoding relevant geometric details. When exploring unknown environments, it needs to be updatable incrementally in an online fashion. We introduce \ourmethod{}, a new method that represents the environment as a Signed Distance Field (SDF). State of the art representations of SDFs are based on either neural networks or voxel grids. Neural networks are capable of representing the SDF continuously. However, they are hard to update incrementally as neural networks tend to forget previously observed parts of the environment unless an extensive sensor history is stored for training. Voxel-based representations do not have this problem but they are not space-efficient especially in large environments with fine details. \ourmethod{} combines the advantages of these representations using a hierarchical approach which employs a coarse voxel grid that captures the observed parts of the environment together with high-resolution local information to train a neural network. \ourmethod{} achieves a 46\% lower mean global SDF error across all test scenes than a state of the art continuous representation, and a 30\% lower error than a discrete representation at the same resolution as our coarse global SDF grid. Videos and code are available at: \url{https://samsunglabs.github.io/HIO-SDF-project-page/}




\end{abstract}


\section{Introduction}


Constructing a geometric representation of a robot's environment which is suitable for real-time planning and control is a fundamental robotics challenge. A good representation must be accurate, support fast inference and, for exploration tasks, incremental update. Signed Distance Field (SDF)~\cite{Curless_SDF_1996} is a scalar field that maps each point in the 3D workspace to its signed distance to the closest obstacle. Representing the robot's environment as a SDF can be an efficient method for performing collision cost queries during reactive planning~\cite{KoptevFigueroaBillard_NeuralImplicit_2023, Vasilopoulos_RAMP_2023}. However, the underlying field must be incrementally updated online.


Discrete online SDF representations such as Voxblox~\cite{Oleynikova_Voxblox_2017}, FIESTA~\cite{Han_FIESTA_2019} or Voxfield~\cite{Pan_Voxfield_2022} have become popular due to their accuracy, relatively low memory footprint and fast update rates. However, discrete methods have several disadvantages. Interpolation is necessary to obtain continuous SDF values and their real-time performance degrades significantly as their voxel resolution increases. Also, because they rely on voxel hashing, they cannot be easily parallelized and therefore are not directly applicable within modern reactive planning schemes such as Model Predictive Path Integral (MPPI)~\cite{Williams_MPPI_2018}, that typically need to make thousands of queries to the SDF module at each iteration. Additionally, such methods only construct the SDF in observed areas and cannot make predictions in occluded regions.

Real-time neural implicit representations such as iSDF \cite{Ortiz_iSDF_2022} offer a more compact description of the underlying SDF, which can be queried for both values and gradients at an arbitrary resolution. However, neural representations also raise technical difficulties. For example, in order to avoid forgetting they typically need to save and replay data from previously visited locations and the learned model tends to over-smooth values resulting in a very coarse representation of the environment. Additionally, due to the absence of a model that can intrinsically capture fine details in the environment, iSDF needs to use heuristics in the definition of the loss function for solving the challenging problem of accurately estimating global Euclidean SDF values from local projective distance estimates for training. Finally, training on a fixed number of sparse data points from the past can lose its efficiency as the space becomes larger.

\begin{figure}[t]
\centering
\includegraphics[width=0.98\columnwidth]{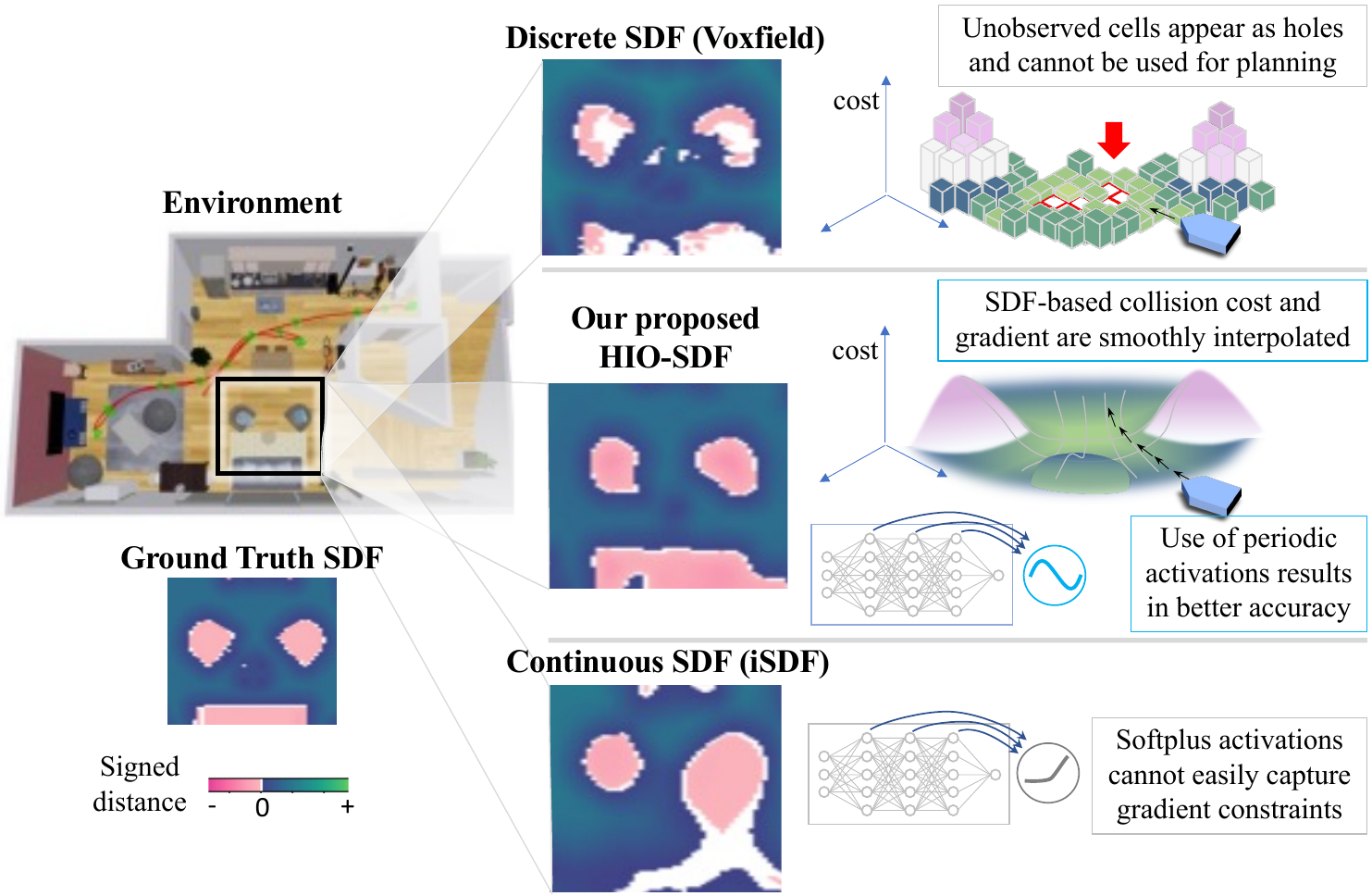}
\caption{\textbf{Overview of approach}: Online discrete Signed Distance Field (SDF) representations can be very accurate but do not include any information in unobserved areas. Existing continuous SDF representations solve that problem but cannot easily capture fine details in the environment. Our proposed method (\ourmethod{}) uses a hierarchy of discrete and continuous algorithms for accurate online 3D SDF reconstruction by a global neural network using periodic activation functions.}
\label{fig:front-figure}
\vspace{-2mm}
\end{figure}

This work seeks to reconcile these two real-time approaches: we propose a two-level hierarchy which contains a global SDF and a local SDF. At the fine level, the local SDF works by performing direct distance computations of query points to the observed point cloud on a GPU, and informs the global SDF updates. The global SDF is built by combining a coarse voxel-based representation (which prevents forgetting) with the local SDF, in order to fit a frequency-based neural network (SIREN~\cite{Sitzmann_SIREN_2020}) to the whole environment.

\subsection{Outline of Contributions}
This paper proposes Hierarchical Incremental Online SDF (\ourmethod{}), a method for constructing a differentiable global Signed Distance Field suitable for real-time application in unknown environments (Fig.~\ref{fig:front-figure}).
The main contributions of our work are as follows:
\begin{enumerate}
    \item Our method uses a mixture of approaches to extract data for the self-supervised training of the global SDF: a) a discrete model for coarse global SDF value look-up without replaying past data, and b) a sensor-based local SDF representation that captures more fine details around the collision boundaries. 
    \item The deep neural network representing the global SDF is constructed using periodic activation functions (SIREN~\cite{Sitzmann_SIREN_2020}), in order to enforce better accuracy and satisfaction of SDF-related constraints. Unlike past work, the use of SIREN allows us to achieve superior performance with a very simple definition of the loss function.
    \item Unlike prior work~\cite{Ortiz_iSDF_2022}, our method offers more flexibility, as it works both with depth sensors and sparse point cloud inputs (e.g., from a 3D LiDAR sensor).
\end{enumerate}

We compare our method against both a discrete SDF representation and a state-of-the-art differentiable SDF model on scenes from two different open-source datasets and a real dataset collected in an office environment. 
The experimental results show that our method constructs a
better underlying SDF representation, even in the presence
of sensor noise and odometry uncertainties, both with dense
depth and sparse 3D LiDAR point cloud data.


\begin{figure}[t]
\centering
\includegraphics[width=1.0\columnwidth]{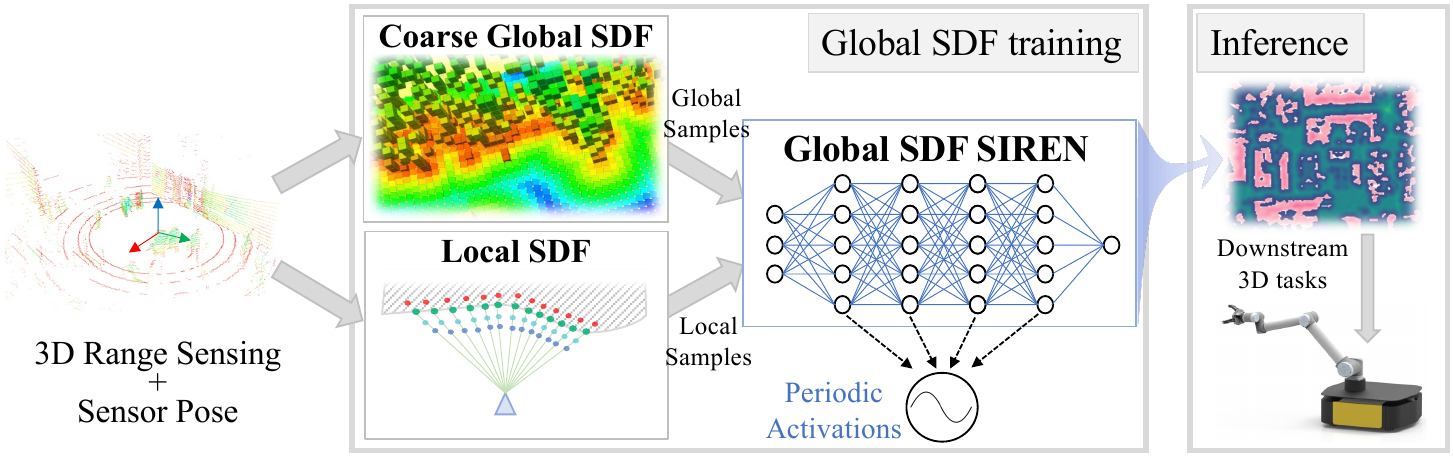}
\caption{\textbf{Overview of our hierarchical incremental online global SDF architecture (\ourmethod)}: The input to the algorithm is a stream of posed point cloud measurements (dense or sparse) and the output is a global implicit 3D SDF map of the environment that can be used directly for downstream tasks. At each time step, the input is consumed by two different modules that produce SDF data (i.e., tuples of 3D coordinates and associated SDF values) for the supervision of a SIREN~\cite{Sitzmann_SIREN_2020} representing our global SDF: 1) a discrete global SDF module fusing measurements in the background at a coarse resolution, and 2) a continuous local SDF module that does not need training and performs brute-force distance computations on the GPU between stratified raycasted samples and the currently observed point cloud. Training of the global SIREN is incremental and online, as new sensor measurements are received.}
\label{fig:architecture}
\vspace{-2mm}
\end{figure}


\section{Related Work}
\label{sec:related-work}

Although occupancy-based representations have been the mainstream in robotics for decades, SDF-based representations have also been widely employed across different domains. SDF was initially suggested for its accuracy and robustness in surface modeling~\cite{Curless_SDF_1996}. Since then, use of SDF for scene reconstruction appear in many previous works~\cite{Carr_RadialBasisFunctions_2001, Newcombe_KinectFusion_2011,Bylow_tracking_2013,Zeng_3DMatch_2017, Yariv_BakedSDF_2023}. Besides its high accuracy as a surface representation, SDF provides rich spatial information of traversable space and thus it is very useful for collision checks and trajectory optimization \cite{Ratliff_CHOMP_2009,Oleynikova_SDF_2016rss}.

Some of the most successful SDF mapping methods in robotics such as Voxblox \cite{Oleynikova_Voxblox_2017}, FIESTA \cite{Han_FIESTA_2019}, and Voxfield \cite{Pan_Voxfield_2022} are based on voxel grids. Voxblox incrementally builds Euclidean Signed Distance Fields (ESDF) from Truncated Signed Distance Fields (TSDF) integrated over time. FIESTA offers improved speed over Voxblox by utilizing occupancy-based wavefront propagation instead of TSDF's projective distance, but at the cost of the sign of the distance field; it builds unsigned distance field maps. Voxfield has improved accuracy over both methods by using non-projective TSDF and surface normal estimates from the incoming point cloud. All of the aforementioned methods run in real-time on a single CPU core and have been released as open source, contributing much to the robotics community. However, although conceptually serving as scalar functions, discrete methods are essentially lookup data structures with limited interpolation and extrapolation capabilities, and make it hard to exploit GPU acceleration for massive queries. 

Alternative studies have focused on continuous SDF representations. Whereas non-parametric regression methods~\cite{Lee_GPIS_2019, Wu_LogGPIS_2023} suffer from increased computational complexity, parametric methods using Deep Neural Networks (DNN) provide a more practical solution in the form of neural implicit functions. One of the earliest methods following this approach is DeepSDF \cite{Park_DeepSDF_2019}, which inspired other studies followed \cite{Gropp_ImplicitGeometricRegularization_2020, Chibane_UnsignedDistanceFields_2020, Sitzmann_SIREN_2020}. While most NN-based SDFs are suggested for offline training schemes, there has been prior work on SDF models for incremental training from online observations. The method in \cite{Yan_Continual_2021} was the first incremental SDF NN algorithm using a relay buffer but did not achieve real-time performance. iSDF \cite{Ortiz_iSDF_2022} is the first real-time incremental continuous SDF. DI-Fusion \cite{Huang_DIFusion_2021} suggests an online method to train SDF using partitioning. Their SDF implementation consists of multiple independent NNs representing local cells, which improves scalability yet sacrifices continuity. Note that these online methods are developed for dense depth sensing and demonstrated in room-scale surface reconstruction applications.

In this study, we seek a solution to incrementally build a continuous SDF for downstream robotic tasks. Prior work has already used SDF representations for offline path refinement~\cite{Ratliff_CHOMP_2009}, learning models for task planning~\cite{Driess_FunctionalsOfSDFs_2021}, predicting collision clearance in high-dimensional configuration spaces \cite{ChaseKew_NeuralCollisionClearance_2021}, or querying the robot's distance to the environment during reactive manipulation planning~\cite{KoptevFigueroaBillard_NeuralImplicit_2023, Liu_DeepSignedDistanceFields_2022, Li_RobotDistanceFields_2023}. 
Our prior work~\cite{Vasilopoulos_RAMP_2023} uses an SDF module that computes raw distances on the GPU between query points and the scene's point cloud for reactive manipulation planning in tabletop scenarios. 
However, this brute-force approach becomes memory inefficient outside of tabletop scenarios as the point cloud of the scene grows.

\section{Problem Statement}
\label{sec:problem-statement}

A 3D SDF parameterized by $\theta$ is defined as an implicit scalar field $f_\theta : \mathbb{R}^3 \mapsto \mathbb{R}$ mapping each 3D location $\mathbf{x} \in \mathbb{R}^3$ to its signed distance $s \in \mathbb{R}$ to the closest surface in the environment, i.e., $f_\theta(\mathbf{x}) = s$. 
The surface of all objects in the environment is implicitly described by the zero-level set of the SDF, $\mathcal{S} := \{ \mathbf{x} \in \mathbb{R}^3 \, | \, f_\theta(\mathbf{x}) = 0\}$.

One important property of a valid SDF is that its gradient at every location $\mathbf{x} \in \mathbb{R}^3$ needs to have unit norm: $|| \nabla f_\theta(\mathbf{x}) || = 1$. Intuitively, this property gives the SDF distance-like properties; it implies that approaching or moving away from the closest surface point of $\mathbf{x}$ by $dx$ in the direction of $\nabla f_\theta(\mathbf{x})$ should decrease or increase the SDF value by the same amount, respectively. This constraint is known as the {\it Eikonal equation}. The gradient $\nabla f_\theta(\mathbf{x})$ of the SDF at a surface point $\mathbf{x} \in \mathcal{S}$ gives the surface normal at this location.

The goal of this work is to construct a continuous and differentiable 3D SDF representation of the environment in real-time from a stream of posed point cloud inputs which is not restricted to dense depth streams unlike prior work~\cite{Ortiz_iSDF_2022}.


\section{Method and System Architecture}
\label{sec:approach}

This section presents \ourmethod{}, our method for estimating global SDF values
(Fig.~\ref{fig:architecture}). The global SDF is represented as a neural network trained in an incremental and online fashion, using data from two separate sources: 1) a discrete global representation running in the background at a coarse resolution, and 2) a brute-force SDF model that computes signed distances of points along rays in the sensor's local field of view to the currently observed point cloud. 

\subsection{Global SDF Network}
\label{subsec:global-sdf}

The global SDF network is a SIREN~\cite{Sitzmann_SIREN_2020} fully-connected neural network with 4 hidden layers of 256 units each, periodic (sinusoidal) activations and a linear output layer. We choose to use a SIREN because it naturally captures the underlying signal from the environment than networks with traditional activation functions (e.g., ReLUs~\cite{Ortiz_iSDF_2022}); the learned weights and biases of the SIREN essentially represent angular frequencies and phase offsets, respectively. An additional property of such networks that make them particularly well-suited for SDF constructions is that the derivative of a SIREN is itself a phase-shifted, well-behaved SIREN, which enables more accurate supervision of the Eikonal constraint that involves the gradient of the network.

Unlike iSDF~\cite{Ortiz_iSDF_2022} and other prior work~\cite{Mildenhall_NeRF_2020, Barron_MipNeRF_2021}, we do not need to apply positional embedding to the 3D input coordinates before feeding them to the network, even though we do not know the scale of the scene ahead of time. The frequency properties of the SIREN directly handle this issue, as discussed above. Additionally, we find that we do not need to feedforward the input to downstream layers~\cite{Ortiz_iSDF_2022}, or introduce any other heuristics to the network for accelerated learning. These are examples that demonstrate the representation power of the SIREN, which allows us to keep a simple architecture. We initialize each layer of the SIREN by drawing weights from the uniform distribution $\mathcal{U}(-\sqrt{6 / n} / \omega_0, \sqrt{6 / n} / \omega_0)$ with $n$ the dimension of the input, as described in the original paper~\cite{Sitzmann_SIREN_2020}. We found that a nominal frequency of $\omega_0 = 10$ works well for our experiments.


\subsection{Coarse Global SDF Data Generation}
Even though any discrete global SDF representation from the literature could be used, we choose Voxfield~\cite{Pan_Voxfield_2022} as the backbone method for generating online coarse global SDF data. Voxfield is shown to outperform other prior work in terms of both accuracy and update time.

At each iteration, Voxfield consumes the most recent posed point cloud input and updates an SDF map of the environment, saved in the form of a voxel grid of a user-specified resolution. Considering that Voxfield's update time and memory consumption increase as the voxel size becomes smaller and/or the environment grows, we use a coarse voxel size (10cm). After each Voxfield update, we sample points and their associated SDF values from the voxel grid; specifically, we sample uniformly at random $N = 10000$ points close to the estimated surface (i.e., points with SDF values less than a fixed threshold $\varepsilon = 5cm$), and $M = 30000$ points away from the surface.

\subsection{Local SDF Data Generation}
\label{subsec:local-sdf}

\begin{figure}[t]
\centering
\includegraphics[width=0.76\columnwidth]{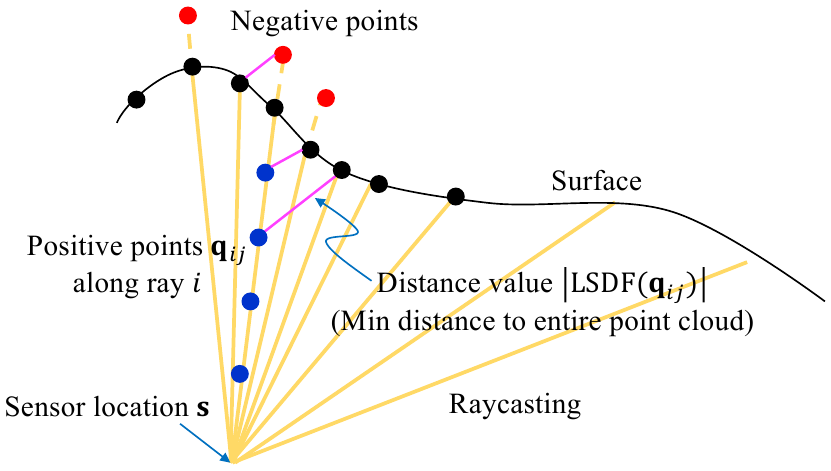}
\caption{\textbf{Local SDF data generation method}: The local SDF module uses raycasting from the sensor location to find points with positive (blue) and negative (red) SDF values before and after the currently observed point cloud (black), respectively. The absolute local SDF value of each point is the minimum distance (magenta) to the entire local point cloud.}
\label{fig:local-sdf}
\vspace{-2mm}
\end{figure}

We complement coarse global SDF data with data generated by a continuous local SDF framework. The local SDF performs direct computation of distances between query points and the currently observed point cloud, in a parallelized fashion on a GPU. This is illustrated in Fig.~\ref{fig:local-sdf}. 
Specifically, given a local point cloud with $P$ points, described as $\mathcal{P} := \{\mathbf{p}_i \in \mathbb{R}^3\}_{i=1,\ldots,P}$, we first sample $S$ points to construct a subset $\mathcal{P}_S \subseteq \mathcal{P}$. Given the current sensor location $\mathbf{s} \in \mathbb{R}^3$, we perform raycasting to find $Q$ stratified samples $\{\mathbf{q}_{ij} \in \mathbb{R}^3\}_{j = 1, \ldots, Q}$ along the rays connecting $\mathbf{s}$ to each point $\mathbf{p}_i \in \mathcal{P}_S$ up to a maximum distance of $(||\mathbf{p}_i - \mathbf{s}|| + t)$ behind the point cloud, with $t \in \mathbb{R}$ the {\it truncation distance}. To each such point $\mathbf{q}_{ij}$, we assign a sign based on its distance to the sensor $\mathbf{s}$ as follows:
\begin{equation}
  \text{sign}(\mathbf{q}_{ij}) :=
    \begin{cases}
      1, & \text{if $||\mathbf{q}_{ij} - \mathbf{s}|| < ||\mathbf{p}_i - \mathbf{s}||$} \\
      -1, & \text{otherwise}
    \end{cases}
\end{equation}

After the computation of these stratified samples with raycasting, we compute the {\it local SDF value} ($\text{LSDF}$) of each sample $\mathbf{q}_{ij}$ as the direct signed distance to the entire local point cloud $\mathcal{P}$:
\begin{equation}
\text{LSDF}(\mathbf{q}_{ij}) := \text{sign}(\mathbf{q}_{ij}) \cdot \min_{k \in {1, \ldots, P}} ||\mathbf{q}_{ij}-\mathbf{p}_k||
\end{equation}
and add all pairs $\{(\mathbf{q}_{ij}, \text{LSDF}(\mathbf{q}_{ij})), \forall i, j\}$ to the dataset for the next global SDF training iteration. Even though this direct computation seems expensive, such distance queries between points in $\mathbb{R}^3$ and the point cloud can be easily parallelized on a GPU, as we demonstrated in our prior work~\cite{Vasilopoulos_RAMP_2023}. In our experiments, we choose $S = 1000$, $Q = 20$ (for a maximum of 20000 local samples) and $t = 0.2m$. We typically perform distance queries between stratified samples and points in the local point cloud in less than 5ms. Using a single point cloud measurement, local SDF values of points away from the surface might not accurately reflect their global SDF values. Therefore, we only keep samples with SDF values less than the truncation distance $t$. This captures fine details around the surface, i.e., the zero level set of the underlying SDF.

\subsection{Loss Function and Training}
\label{subsec:loss-training}

Each time a new posed point cloud measurement is received, we run a Voxfield update step and concatenate the coarse global SDF data with the local SDF data. Using this data, consisting of pairs $(\mathbf{x}, s)$ of 3D coordinates and their associated SDF values, we run a training iteration of the global SDF network to update its weights $\theta$. For supervision, we use the following loss function
\begin{equation}
    \mathcal{L}_{\text{total}}(\theta) := \lambda_{\text{SDF}} \, \mathcal{L}_{\text{SDF}}(\theta) + \lambda_{\text{Eikonal}} \, \mathcal{L}_{\text{Eikonal}}(\theta)
\end{equation}
with the SDF loss defined as the $L_1$ norm
\begin{equation}
    \mathcal{L}_{\text{SDF}}(\theta) := \mathcal{L}_{\text{SDF}}(f_\theta(\mathbf{x}), s) = |f_\theta(\mathbf{x}) - s| \label{eq:sdf-loss}
\end{equation}
and the Eikonal loss defined as
\begin{equation}
    \mathcal{L}_{\text{Eikonal}}(\theta) := \mathcal{L}_{\text{Eikonal}}(\nabla f_\theta(\mathbf{x})) = \left| ||\nabla f_\theta(\mathbf{x})|| - 1\right|
    \label{eq:eikonal-loss}
\end{equation}
For our experiments, we set $\lambda_{\text{SDF}} = 5.0$, $\lambda_{\text{Eikonal}} = 2.0$, and train for 10 epochs per incremental update, using batch optimization and the Adam optimizer with learning rate 0.0004 and weight decay 0.012.

Our loss function is straightforward to implement and does not require an elaborate SDF loss definition, which often depends on the distance to the surface, the truncation of the Eikonal loss for samples close to the surface, or direct supervision on the direction of SDF gradients. These introduce either additional hyper-parameters or additional pre-processing of the data (e.g., the requirement for surface normals estimation~\cite{Ortiz_iSDF_2022}, which might not be an option with sparse point cloud data from a 3D LiDAR).
\section{Results}
\label{sec:results}

In this section, we compare \ourmethod{}  against  two state-of-the-art approaches: iSDF and Voxfield. Note that Voxfield also provides our coarse global SDF estimates, but we will see that our method outperforms Voxfield at 10cm resolution when using a 10cm resolution grid and is even competitive with Voxfield at 5cm resolution. In our experiments, three datasets are used: ReplicaCAD~\cite{Szot_ReplicaCAD_2021}, ScanNet~\cite{Dai_ScanNet_2017}, and data collected from our office which we call `SAIC-NY'. For ReplicaCAD and ScanNet, we use similar scenes as those used in iSDF (\texttt{apt\_2\_nav}, \texttt{apt\_2\_obj}, \texttt{apt\_2\_mnp}, \texttt{apt\_3\_nav} from ReplicaCAD and \texttt{scene0030\_00}, \texttt{scene0004\_00} from ScanNet) and follow the same evaluation method. The SAIC-NY dataset includes depth images from an Azure Kinect camera and Velodyne 3D LiDAR scans collected by a Clearpath Ridgeback mobile robot. We use this dataset to compare all methods in a large unmapped space with uncertain odometry (computed online using gMapping~\cite{Grisetti_gMapping_2007}), as well as demonstrate \ourmethod{}'s flexibility with different sensor inputs.

For the evaluation of \ourmethod{}, we play back data in ROS and let the whole pipeline run online, training for 10 epochs per incremental update of the global SDF. The data playback is not paused when evaluating our network, in order to also test the online capabilities of our approach. The reader is referred to the accompanying video submission for more experiments and a demonstration of our algorithm.

\subsection{Qualitative Evaluation} 
\begin{figure*}[t]
\centering
\includegraphics[width=0.95\textwidth]
{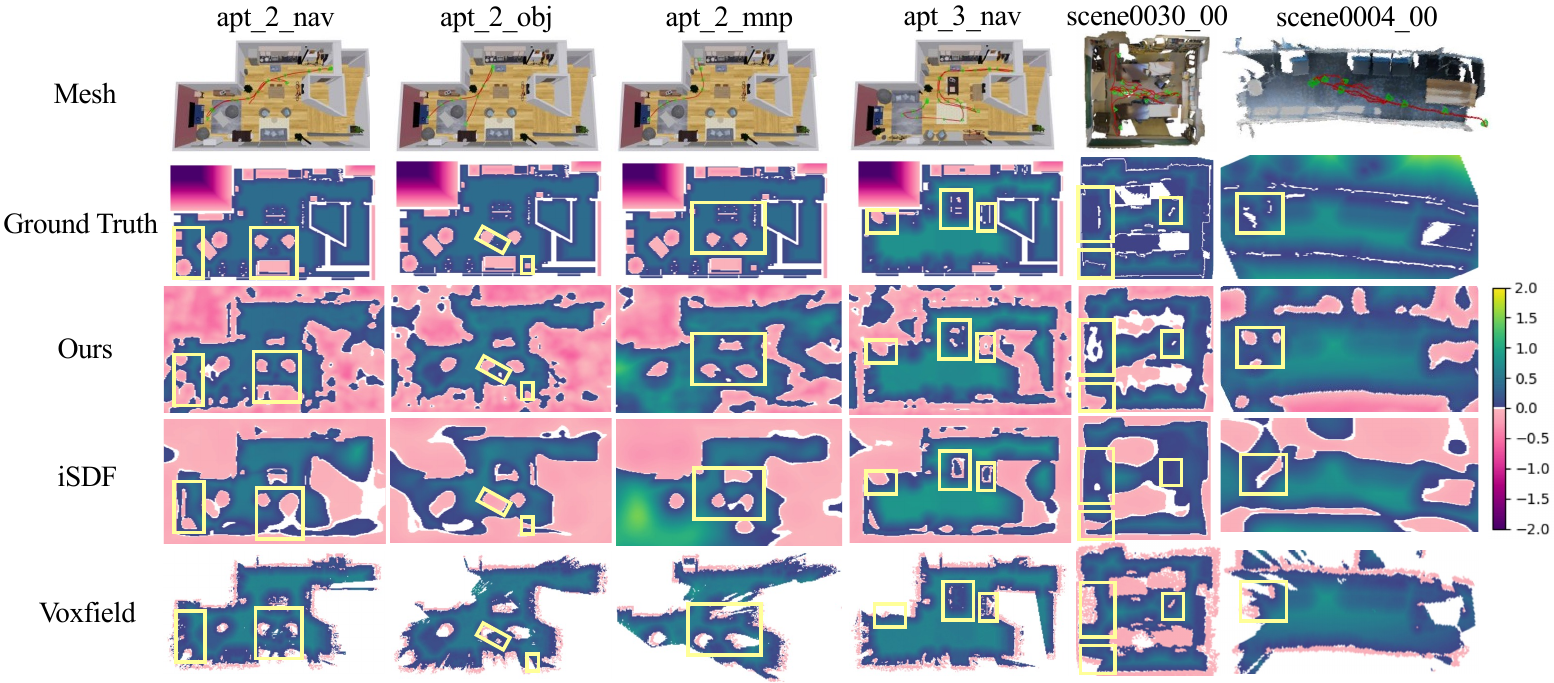}
\caption{\textbf{Global SDF slice comparison}: The sensor trajectories for each scene are shown in red at the top row with ground truth meshes. SDF slices are computed using the final SDF at the end of each sequence. \ourmethod{} is able to represent more detailed SDF information around furniture, compared to both iSDF and Voxfield. For Voxfield, invisible regions by the sensor are shown in white. Details captured by our method even though not captured by the underlying coarse-SDF Voxfield backbone are highlighted in yellow.}
\label{fig:sdf-slices}
\vspace{-2mm}
\end{figure*}
\begin{figure*}[t]
\centering
\includegraphics[width=0.94\textwidth]{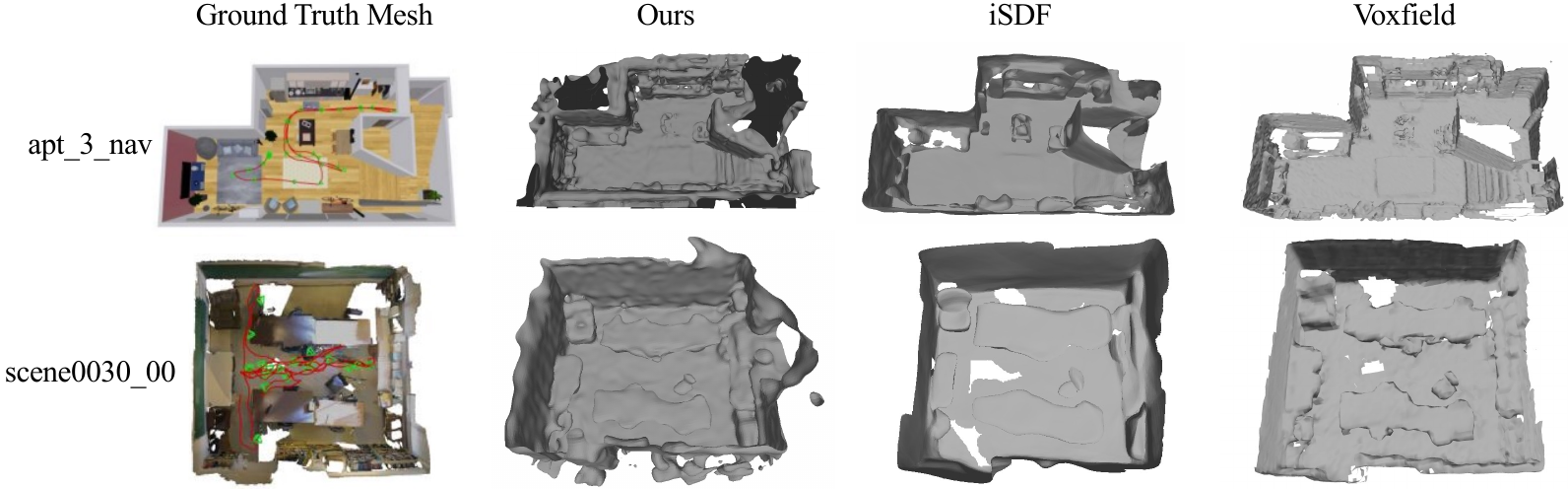}
\caption{\textbf{Mesh reconstruction comparison}: By virtue of using SIREN and utilizing both continuous local and coarse global information, \ourmethod{} generates complete and smooth surfaces that also include finer geometric details over the entire environment compared to iSDF and Voxfield.}
\label{fig:meshes}
\vspace{-2mm}
\end{figure*}
\begin{figure*}
\centering
\includegraphics[width=0.93\textwidth]{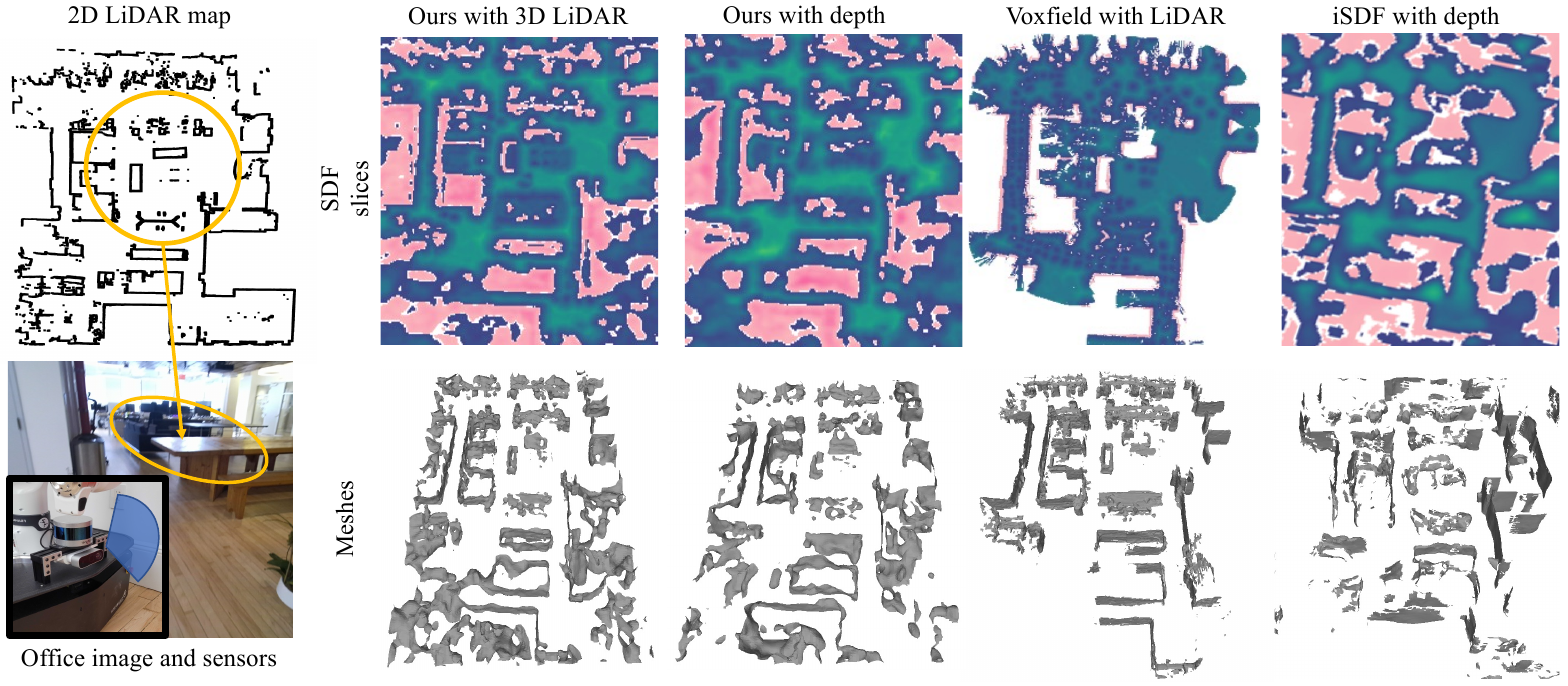}
\caption{\textbf{Qualitative results in the SAIC-NY office dataset collected with a Clearpath Ridgeback}: Unlike iSDF, \ourmethod{} works with both sparse 3D LiDAR data from a Velodyne sensor and dense depth information. Unlike Voxfield, it also attempts to plausibly fill out missing information.}
\label{fig:office}
\vspace{-2mm}
\end{figure*}
Fig.~\ref{fig:sdf-slices} compares global SDF slices of the three methods at a particular height against ground truth slices in six different environments. For a fair comparison, we evaluate the SDF values from all three approaches using a 3D mesh grid of the same resolution and interpolate Voxfield's coarse discrete values at a finer resolution. The used height for all environments is close to the floor where most of the geometric details in the environment lie (tables, chairs, etc.).
As expected and shown in Fig. \ref{fig:sdf-slices}, \ourmethod{} and iSDF can continuously represent the entire space with SDF values, while Voxfield includes areas with nonexistent SDF values. In addition, \ourmethod{} is able to better capture detailed information around furniture compared to iSDF, as highlighted in yellow. 
Fig \ref{fig:meshes} includes a qualitative comparison of \ourmethod{}'s reconstructed 3D meshes against iSDF and Voxfield. The meshes in the figure are generated using the zero level set of SDF values with marching cubes on uniform 3D grids using iSDF's visualization methods.
\ourmethod{} generates smooth surfaces that also include finer geometric details over the entire environment compared to iSDF, while the discrete Voxfield cannot generate a complete mesh.
Overall, our approach qualitatively outperforms the other two approaches as it has less holes and more detail.

\begin{figure*}[t]
\centering
\includegraphics[width=0.92\textwidth]{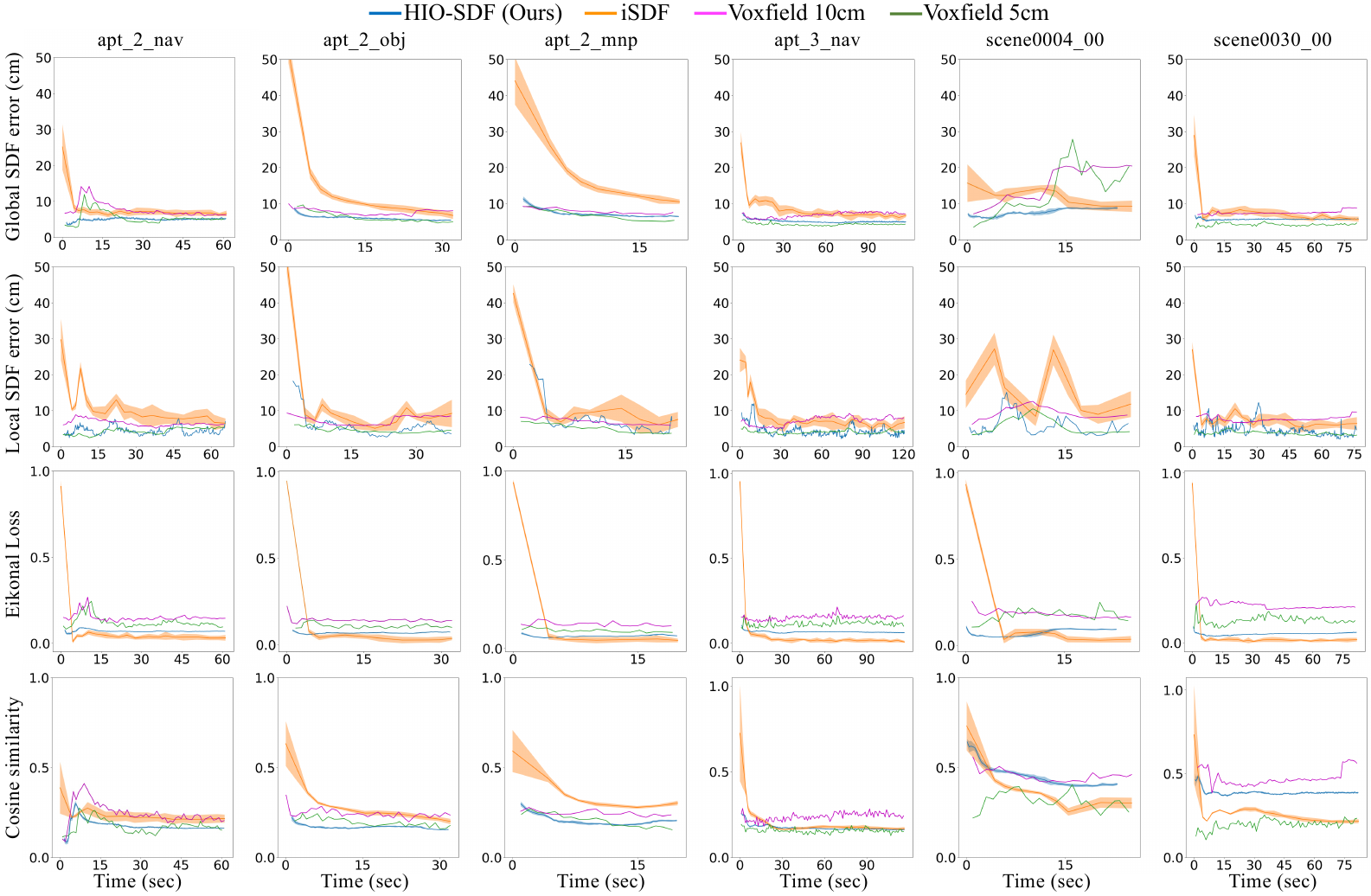}
\caption{Quantitative comparison of all methods across 6 different scenes (4 scenes from ReplicaCAD and 2 scenes from ScanNet). The $x$-axis represents the sequence running time (wall clock) as the simulated sensor navigates each environment.}
\label{fig:graphs}
\vspace{-2mm}
\end{figure*}

\subsection{Quantitative Evaluation}

Fig.~\ref{fig:graphs} includes our quantitative comparisons. We compare \ourmethod{}, iSDF and Voxfield against ground truth data over time (as each sequence progresses), by evaluating SDF values and gradients on 25000 points sampled uniformly at random from the visible region at each time instant. For iSDF, we use their released evaluation method; for our method and Voxfield we evaluate the error each time a new measurement is received. Two resolutions of Voxfield grid are evaluated: one at 10cm, which also serves as the backbone coarse global SDF for \ourmethod{}, and the other at 5cm that is more accurate yet approaches the limits of Voxfield's real-time capabilities. For each scene, we run our method and iSDF 10 times to capture the variance but only once for Voxfield since it is a deterministic algorithm. We evaluate all methods using 4 different metrics against ground truth data: 1) the SDF error defined in Eq.~\ref{eq:sdf-loss} using sampled points from the entire visible region (global SDF error), 2) the same SDF error using only local points within a radius of 3m from the sensor position at each time instant (local SDF error), 3) the Eikonal loss in Eq.~\ref{eq:eikonal-loss}, and 4) the cosine similarity between the predicted gradient $\nabla f_\theta(\mathbf{x})$ and the ground truth SDF gradient $\mathbf{g}$, defined as
\begin{equation*}
    \mathcal{L}_{\text{sim}}(\mathbf{x}) := 1 - \frac{\nabla f_\theta(\mathbf{x}) \cdot \mathbf{g}}{||\nabla f_\theta(\mathbf{x})|| \cdot ||\mathbf{g}||}
\end{equation*}

Our quantitative comparisons present some interesting results. First, looking at the global and local SDF errors, our method is more accurate than iSDF and Voxfield at 10cm resolution and is competitive with Voxfield at 5cm resolution, even though it just uses a 10cm-resolution voxel grid as its discrete global SDF backbone. We achieve a mean global SDF error of 5.57cm across all scenes, compared to iSDF's 10.27cm error and Voxfield's 7.98cm and 5.48cm error at 10cm and 5cm resolution, respectively. Additionally, being trained from this discrete representation, our initial errors are much smaller than iSDF and rapidly converge to their final values, and our measurements have much lower variance across all metrics/scenes. Voxfield errors at both resolutions are much higher than both our method and iSDF in the ScanNet scenes. We believe this happens because ScanNet includes scenes where the measurements exceed the scale of the room, introducing additional noise to the point cloud measurements; in contrast, \ourmethod{} and iSDF are capable of filtering out such noise with neural network interpolation. The Eikonal loss error for each method is fairly constant across different sequences, with iSDF performing slightly better than \ourmethod{}. Lastly, the value of the synergy between the continuous local and coarse global data is captured in our cosine similarity plots: even though iSDF explicitly supervises this loss to achieve good performance, our method achieves better results (sometimes by a wide margin) in all ReplicaCAD scenes and the loss still decreases or stays constant in ScanNet scenes, implying that the global SIREN SDF is also capable of accurately capturing the direction of the SDF gradient without direct supervision.

\subsection{Evaluation in SAIC-NY Office Dataset}

Fig.~\ref{fig:office} includes a comparison of SDF reconstruction results in our independently-collected SAIC-NY office dataset. \ourmethod{} is able to construct high-quality SDF maps, even in the presence of uncertain odometry measurements. We use this dataset, which represents a large 25m$\times$20m office space, to also keep track of timings and memory consumption. By the end of the run, \ourmethod{} and Voxfield needed to store a 42.6MB coarse global grid. By comparison, iSDF needed to store 178 depth keyframes of 1.8MB each, accounting for 320.4MB in total. The memory needed for storing the weights of both \ourmethod{}'s and iSDF's network was similar (approximately 1MB). In terms of update times, by the end of the run our method requires around 160ms to perform a Voxfield update, 270ms to extract global SDF data, 38ms to construct the local SDF data, and 29ms per training epoch of the global SIREN, for a total update time of 1sec for 10 training epochs. A forward pass and a backward pass through the SIREN take around 1ms and 9ms respectively on a NVIDIA GeForce RTX 2080 Ti GPU, while querying the final trained model with 100,000 3D coordinates requires only 9ms compared to Voxfield's 40ms. Overall, evaluations in this dataset demonstrate \ourmethod{}'s time- and memory-efficiency, as well as its ability to construct accurate SDFs in large environments extending the room-scale.
\section{Conclusion}
\label{sec:conclusion}

This paper presents \ourmethod{}, a hierarchical, incremental and online global Signed Distance Field model, represented by a SIREN neural network with periodic activation functions. Our approach combines the benefits of coarse, discrete global SDF representations and locally-accurate, continuous SDF models. In several runs through a variety of datasets and scenes, we demonstrate that \ourmethod{} is more accurate and efficient than other state-of-the-art representations.
Future work will seek to include applications of \ourmethod{} in mobile manipulation for reactive 3D planning and motion control.

\small
\bibliographystyle{IEEEtran}
\bibliography{references}

\end{document}